\pdfoutput=1

\documentclass[11pt]{article}

\usepackage[final]{acl} 
\usepackage{times}      
\usepackage{latexsym}
\usepackage[T1]{fontenc}
\usepackage{microtype}
\usepackage{inconsolata}
\usepackage{amsfonts, amssymb, amsmath, bbm}

\usepackage{booktabs}  
\usepackage{multirow}
\usepackage{multicol}
\usepackage{hhline}
\usepackage{makecell}
\usepackage{array}
\usepackage{tabularx}
\usepackage{graphicx}
\usepackage{subcaption} 
\usepackage{float}
\usepackage{placeins}   
\usepackage{caption}
\usepackage{amsmath}

\usepackage{tikz}
\usetikzlibrary{shapes.geometric, arrows, positioning, matrix}
\usepackage{pgfplots}
\usepackage{pgfplotstable}
\pgfplotsset{compat=1.18}

\usepackage{verbatim}
\usepackage{enumitem}
\usepackage{relsize}
\usepackage{natbib}
\usepackage{listings}
\usepackage{xurl}
\usepackage[disable]{todonotes}

\newcommand{\cmark}{\checkmark}
\newcommand{\xmark}{\texttimes}

\title{ReFACT: A Benchmark for Scientific Confabulation Detection\\with Positional Error Annotations}

\author{
    {\bf Yindong Wang}\textsuperscript{*} \quad
    {\bf Martin Preiß}\textsuperscript{*} \quad
    {\bf Margarita Bugue\~no} \\
    {\bf Jan Vincent Hoffbauer} \quad
    {\bf Abdullatif Ghajar} \quad
    {\bf Tolga Buz} \quad
    {\bf Gerard de Melo}\\
    Hasso Plattner Institute / University of Potsdam, Germany \\
    {\normalsize \texttt{\{yindong.wang,margarita.bugueno,tolga.buz,gerard.demelo\}@hpi.de}} \\
    {\normalsize \texttt{\{martin.preiss, janvincent.hoffbauer, abdullatif.ghajar\}@student.hpi.uni-potsdam.de}}\\
    \textsuperscript{*}Equal contribution
}

\begin{document}
\maketitle

\begin{abstract}
The mechanisms underlying scientific confabulation in Large Language Models (LLMs) remain poorly understood. We introduce \textbf{Reddit False And Correct Texts} (ReFACT), a benchmark of 1,001 expert-annotated question–answer pairs with \textbf{span-level error annotations} derived from Reddit's \textit{r/AskScience}. Evaluating 9 state-of-the-art LLMs reveals two critical limitations. First, models exhibit a dominant \textbf{salient distractor} failure mode: 61\% of incorrect span predictions are semantically unrelated to actual errors. Crucially, this pattern persists across all model scales (1B to 70B), indicating a fundamental semantic grounding deficit that \textbf{scaling alone fails to resolve}. Second, we find that \textbf{comparative judgment} is paradoxically harder than independent detection—even GPT-4o's F$_1$ score drops from 0.67 to 0.53 when comparing answers side-by-side. These findings directly challenge the reliability of LLM-as-Judge paradigms for scientific factuality. Code and data are released at \url{https://github.com/ddz5431/ReFACT}.
\end{abstract}

\begin{figure*}[!ht]
    \centering
    \includegraphics[width=0.95\textwidth]{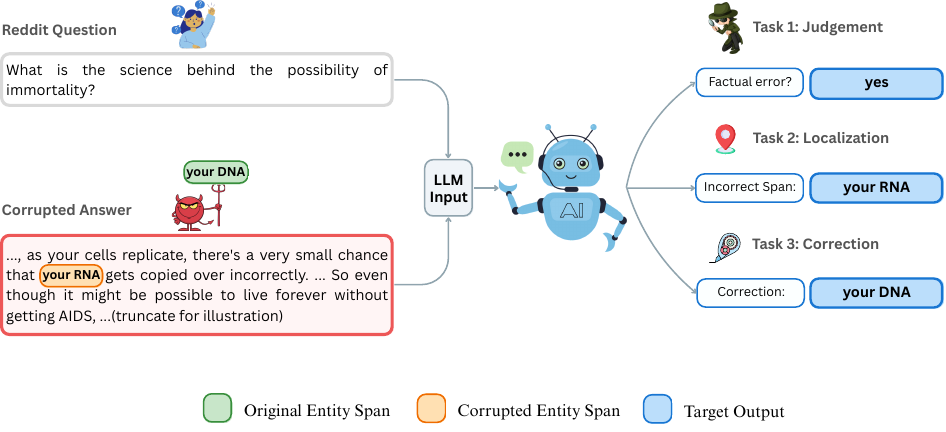}
    \caption{\textbf{Overview of the ReFACT Evaluation Pipeline.} Illustrated via an entity replacement example (``your DNA'' $\rightarrow$ ``your RNA''), the benchmark evaluates three capabilities: (1) \textbf{Judgment} – detecting the confabulation, (2) \textbf{Span Localization} – identifying the corrupted span, and (3) \textbf{Correction} – recovering the original entity.}
    \label{fig:pipeline}
\end{figure*}

\section{Introduction}
While Large Language Models (LLMs) demonstrate impressive fluency, they remain prone to scientific confabulation---generating content that sounds plausible to non-experts but is fundamentally factually incorrect. Unlike obvious factual errors that can be easily verified, scientific misinformation is often subtle and domain-specific. Such inaccuracies may appear credible and require expert knowledge to detect, presenting a critical challenge for reliable knowledge dissemination in online discourse.

We adopt the term \emph{scientific confabulation} to describe fluent, plausible yet factually incorrect scientific text, which are often also informally referred to as hallucinations.
Following clinical usage, confabulation refers to fabricated but coherent content that appears convincing~\citep{sui-etal-2024-confabulation}. Scientific confabulations are particularly insidious due to their surface fluency and domain-specific language, often requiring expert knowledge to detect (e.g., describing DNA replication using RNA mechanisms; see \autoref{fig:pipeline}).

To address this gap in the evaluation landscape, we introduce \textbf{Reddit False And Correct Texts} (ReFACT), the first benchmark specifically designed to evaluate LLMs' ability to detect, localize, and correct confabulations. ReFACT is constructed from authentic, human-authored scientific discourse in r/AskScience—a community with over 23 million members and rigorous moderation standards. Our contributions include:

\begin{itemize}[noitemsep, leftmargin=10pt]
\item \textbf{Scientific Confabulation Benchmark}: 1,001 question--answer pairs across 10 scientific domains with fine-grained span-level confabulation annotations and error types.
\item \textbf{Three-Task Evaluation}: (1) binary confabulation detection judgment, (2) span-level localization, and (3) confabulation correction generation.
\item \textbf{Human-Verified Pipeline}: LLM-assisted corruption of authentic answers with multi-annotator verification ensuring high-quality confabulations.
\item \textbf{The Salient Distractor Phenomenon}: We identify a scale-invariant failure mode where models fixate on contextually salient terms rather than factual errors. This limitation persists from 1B to 70B parameters, providing empirical evidence that model scaling alone is insufficient for scientific factuality.
\end{itemize}

\section{Related Work}
\label{sec:related_research}

\begin{figure*}[!ht]
    \centering
    \includegraphics[width=0.9\textwidth]{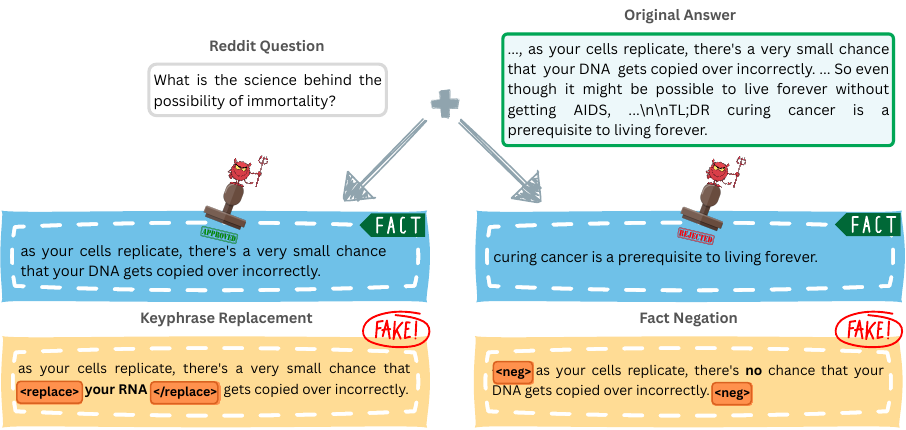}
    \caption{\textbf{Confabulation Generation Pipeline.} Illustrated via the negation and replacement strategies, the process transforms factual claims through: (1) \textbf{Extraction} and \textbf{Selection} of salient facts; (2) \textbf{Transformation} via logical \textit{negation} or domain-specific \textit{entity replacement} (e.g., ``your DNA'' $\rightarrow$ ``your RNA''); and (3) \textbf{Annotation} of precise error spans.}
    \label{fig:transformations}
\end{figure*}

\subsection{Factuality, Hallucination \& Confabulation}
A range of prior work has attempted to define and categorize errors produced by LLMs, but terms such as \emph{hallucination}, \emph{factuality}, and \emph{faithfulness} are often used inconsistently~\citep{ji2023survey}.
\emph{Factuality} typically refers to agreement with external or real-world facts, while \emph{hallucination}, adapted from psychology, describes model outputs that introduce information not grounded in reality. This phenomenon extends beyond simple textual errors~\citep{ji2023survey}, manifesting in multimodal inconsistencies~\citep{wang2025image} or failures in uncertainty calibration~\citep{cohen2024idk}. Conversely, \emph{faithfulness} denotes strict consistency with the source input, independent of real-world factual correctness~\citep{maynez2020faithfulness, augenstein2023factualitychallengeseralarge}.

Recently, the term \emph{confabulation} has gained traction to describe a distinct error category characterized by fluent, coherent, and contextually plausible content that nevertheless contains subtle factual inaccuracies~\citep{sui-etal-2024-confabulation}.
For instance, a model might claim, ``\textit{mitochondria produce glucose through cellular respiration}''---a statement plausible to non-experts but is biologically inaccurate.
While prior work has explored confabulation in general domains, domain-specific epistemic failures remain largely overlooked. To address this gap, we introduce a benchmark specifically designed for fine-grained evaluation of scientific confabulation.

\subsection{Factuality and Hallucination Benchmarks}

Research on misinformation has shifted from external social media cues toward detecting hallucinations intrinsic to LLM-generated text~\citep{chen2023fakeFactRecognitionSurvey}.
Initial benchmarks prioritized \textbf{binary classification}: TruthfulQA~\citep{lin2022truthfulQa} employs a GPT-3 judge, complemented by prompt-based~\citep{lee2023factualityPrompt} and agentic approaches~\citep{li2024llmAgentForFakeNewsDet}.
However, the reliability of LLMs as evaluators remains contested. While some studies suggest LLMs excel at verification~\citep{guan2024lmsHallucinateButMayExcelAtFactVerification,huang2024fakegpt}, \citet{hu2024badActorGoodAdvisor} demonstrate that specialized BERT-based architectures still outperform general-purpose LLMs on binary detection tasks.

To enable fine-grained analysis, recent benchmarks advance toward \textbf{span-level localization}. FactScore~\citep{min2023factsScore} assesses atomized statements via retrieval-based scoring. \textsc{Fava\-Bench}~\citep{mishra2024favaBench} and HALoGEN~\citep{ravichander2025halogenfantasticllmhallucinations} introduce detailed error taxonomies (e.g., contradictions, intent), yet rely on model-generated outputs rather than human authorship. Similarly, Halu\-Eval 2.0~\citep{halueval2} and Hallu\-Lens~\citep{bang2025hallulensllmhallucinationbenchmark} are limited to synthetic or general-domain content, lacking the complexity of authentic scientific discourse.

\begin{table*}[!ht]
\footnotesize
\centering
\begin{tabularx}{\textwidth}{@{}lXlccccc@{}}
\toprule
\textbf{Benchmark} & \textbf{Source} & \textbf{Samples} & \textbf{Length} & \textbf{Binary} & \textbf{Span} & \textbf{Error Type} & \textbf{Human?} \\
\midrule
SimpleQA~\cite{wei2024simpleqa}                                 & Curated questions   & 4,326   & 1--3                    & \cmark     & \xmark    & \xmark  & \cmark      \\
TruthfulQA~\cite{lin2022truthfulQa}                             & Curated questions   & 817  & 10--30                   & \cmark     & \xmark    & \xmark  & \cmark      \\
HaluEval~\cite{li-etal-2023-halueval}                           & Mixed (real/synth)  & 35,000   & \textasciitilde 37      & \cmark     & \xmark    & \xmark  & Mixed       \\
HaluEval 2.0~\cite{halueval2}                                   & Mixed (real/synth)  & 8,770   & \textasciitilde 37      & \cmark     & \xmark    & \xmark  & Mixed       \\
\textsc{Fava\-Bench}~\cite{mishra2024favaBench}                            & Model outputs       & 1,000   & \textasciitilde 30      & \xmark     & \cmark    & \cmark  & Mixed       \\
HALoGEN~\cite{ravichander2025halogenfantasticllmhallucinations} & Model outputs       & 10,923   & \textasciitilde 23      & \xmark     & \cmark    & \cmark  & \xmark      \\
FELM~\cite{chen2023felmbenchmarkingfactualityevaluation}        & Model outputs       & 847  & \textasciitilde 89      & \xmark     & \cmark    & \cmark  & \xmark      \\
HalluLens~\cite{bang2025hallulensllmhallucinationbenchmark}     & Wikipedia           & 5,000   & \textasciitilde 40      & \cmark     & \cmark    & \xmark  & Mixed       \\
\midrule
\textbf{ReFACT}~(This work)   & \textbf{Reddit} & \textbf{1,001} & \textbf{130+} & \cmark     & \cmark    & \cmark & \cmark     \\
\bottomrule
\end{tabularx}
\caption{
\textbf{Benchmark Comparison}. ReFACT is the only benchmark with fully human-verified span-level and error-type annotations on long-form scientific QA. Columns: \textbf{Samples} (dataset size), \textbf{Length} (answer words), \textbf{Binary} (binary labels), \textbf{Span} (span localization), \textbf{Error Type} (error categorization), \textbf{Human} (human verified; \textit{Mixed} = partial).
}
\label{tab:benchmarks}
\end{table*}

\section{ReFACT Creation Methodology}
\label{sec:dataset_creation_pipeline}

Our dataset construction prioritizes \textbf{real-world grounding} over synthetic generation. Unlike benchmarks relying on LLM-generated QA pairs \citep{mishra2024favaBench, ravichander2025halogenfantasticllmhallucinations}, we source factual content from Reddit's r/AskScience to capture community-validated scientific discourse. This ensures a verifiable factual basis, while subsequent transformations introduce subtle confabulations that preserve contextual plausibility. Such a design is critical for evaluating a model's ability to identify nuanced factual inaccuracies rather than the obvious artifacts typical of synthetic data.

\subsection{Data Collection and Quality Control}
We source data from r/AskScience~\footnote{\url{https://www.reddit.com/r/AskScience}}, a strictly moderated community dedicated to scientific inquiry. While upvotes generally correlate with quality, social dynamics (e.g., humor) can introduce noise.

\paragraph{Filtering} To mitigate this, we select only top-rated questions (score $\geq$ 4) and their best-scoring answers. We further retain pairs with 500--1,000 characters to avoid non-descriptive or overly verbose content, yielding 10,282 pairs (statistics in \autoref{sec:dataset-statistics}).

\subsection{Confabulation Generation}
\label{subsec:confab_gen}
We systematically corrupt factual answers using two strategies: \textbf{negation} ( logical reversal) and \textbf{entity replacement} (term substitution). To ensure the generated confabulations remain fluent and contextual plausible while being factually incorrect, we employ \texttt{Gemma-2-27B-it}~\cite{gemmateam2024gemma2improvingopen} within a multi-stage prompt pipeline (see \autoref{sec:transformation_prompts} and \autoref{fig:transformations}). Crucially, for entity replacement, we integrate an NLI classifier~\cite{laurer_less_2022} to filter out synonyms substitutions, guranteeing that the generated entities represent distinct semantic errors rather than mere paraphrases.

\subsection{Human Annotation Procedure}
\begin{sloppypar}
Using the \texttt{doccano} platform \cite{doccano}, annotators evaluated each triplet $(q_i, a_i, t(a_i))$ against strict validity criteria. To be retained, a transformed answer $a_i$ must be: (1) \textbf{coherent} and contextually relevant; (2) \textbf{factually} incorrect relative to scientific consensus; and (3) \textbf{precisely tagged} (e.g., \texttt{<replace>}). This rigorous process filters out sarcasm or low-quality generations, ensuring only subtle, plausible confabulations are preserved. Detailed guidelines are available in our repository~\footnote{\url{https://github.com/ddz5431/ReFACT}}. 
\end{sloppypar}

\subsection{Results}

\paragraph{Transformation Success Rate}
Defined as the proportion of generations passing human validation for plausibility and falsity, our pipeline achieved success rates of 58\% for negation and 57\% for replacement in the final batch ($N=400$). The released dataset comprises exclusively these strictly verified samples.

\paragraph{Annotator Agreement}
To assess the reliability of our validity criteria, we measured pairwise inter-annotator agreement among three annotators on a random subset of 100 samples in which all three annotators overlapped.
The resulting agreement rate is $72.56\%$, suggesting consistent application of the annotation guidelines.

\paragraph{Resulting Dataset}
The final dataset comprises 1,001 samples, balanced between negation ($N=527$) and entity replacement ($N=474$). Each entry consists of the original question--answer pairs aligned with its. Further statistics on sample length and the distribution of knowledge domains can be found in \autoref{sec:dataset-statistics}.

\section{Benchmarking LLMs with ReFACT}
\label{sec:experiments}

\subsection{Models and Implementation}
We evaluate instruction fine-tuned versions of \texttt{Gemma-3} (1B, 4B, 12B, 27B)~\citep{gemmateam2025gemma3technicalreport}  and Llama-3 (1B, 3B, 70B)~\citep{grattafiori2024llama3herdmodels} alongside GPT-4o~\citep{openai2024gpt4technicalreport} as a proprietary reference. All experiments exmploy zero-shot prompting with task-specific templates (details in \autoref{sec:experiment_prompts}).

\subsection{Evaluation Tasks and Metrics}
We propose a structured evaluation framework to assess LLM's capabilities in handling scientific confabulation across three key dimensions: Judgment, Localization, and Correction. \autoref{fig:pipeline} illustrates the evaluation pipeline using an entity replacement example.

\paragraph{Task 1: Confabulation Judgment}~\\
\textbf{Input}: A pair $(q, a)$ (Independent) or a triplet $(q, a_1, a_2)$ (Comparative). \newline
\textbf{Output}: A predicted character span $\hat{s}$ identifying the error (entity-level for replacement, sentence-level for negation). \newline
\textbf{Metrics:} We report Accuracy and F$_1$ score for the confabulated class.

\paragraph{Task 2: Confabulation Localization}~\\
\textbf{Input:} A question $q$ and a confabulated answer $a_{\mathrm{conf}}$.\newline
\textbf{Output:} A predicted character span $\hat{s}$ identifying the error (entity-level for replacement, sentence-level for negation). \newline
\textbf{Metrics:} We compute the Intersection-over-Union (IoU) between the predicted span $\hat{s}$ and the gold span $s^*$. Localization is considered accurate if the overlap exceeds 50\%:\newline
\begin{equation}
\mathrm{Acc}_{\mathrm{loc}} = \frac{1}{N}\sum_{i=1}^{N}\mathbbm{1}\left[ \frac{\lvert \hat{s}_i \cap s_i^* \rvert}{\lvert \hat{s}_i \cup s_i^* \rvert} \ge 0.5 \right]
\end{equation}

\paragraph{Task 3: Entity Correction}~\\
\textbf{Input:} a tuple $(q, a_{\mathrm{conf}})$, where the error span in $a_{\mathrm{conf}}$ is explicitly marked (e.g., via \texttt{<replace>}). \newline
\textbf{Output:} a corrected entity string $\hat{e}$ \newline
\textbf{Metrics:} Exact Match (EM) accuracy on whitespace-normalized strings:
\begin{equation}
\mathrm{Acc}_{\mathrm{corr}} = \frac{1}{N}\sum_{i=1}^{N} \mathbbm{1}[\mathrm{norm}(\hat{e}_i) = \mathrm{norm}(e_i^*)]
\end{equation}
We exclusively evaluate entity replacement, as preliminary experiments indicated that reversing logical negation is a trivial task for current models (near-perfect performance).

\paragraph{Aggregate Performance.}
We report the macro-average accuracy ($\mathrm{Acc}_{\mathrm{avg}}$) across the five sub-tasks: Independent/Comparative Judgment ($\mathrm{J}_{\mathrm{ind}}, \mathrm{J}_{\mathrm{comp}}$), Negation/Entity Localization ($\mathrm{L}_{\mathrm{neg}}, \mathrm{L}_{\mathrm{ent}}$), and Entity Correction ($\mathrm{C}_{\mathrm{ent}}$):

\begin{equation}
    \mathrm{Acc}_{\mathrm{avg}} = \frac{1}{5} \left(
    \begin{aligned}
        &\mathrm{Acc}_{\mathrm{J}}^{\mathrm{ind}} + \mathrm{Acc}_{\mathrm{J}}^{\mathrm{comp}} \\
        + \: &\mathrm{Acc}_{\mathrm{L}}^{\mathrm{neg}} + \mathrm{Acc}_{\mathrm{L}}^{\mathrm{ent}} \\
        + \: &\mathrm{Acc}_{\mathrm{C}}^{\mathrm{ent}}
    \end{aligned}
    \right)
\end{equation}

\paragraph{Metric Selection.}
For \textbf{Judgment}, we report F$_1$ to balance precision and recall given the class imbalance.
For \textbf{Localization}, we adopt the standard Intersection-over-Union metric (IoU $\ge$ 0.5) following extractive QA benchmark\citep{rajpurkar-etal-2016-squad}.\linebreak
Crucially, for \textbf{Correction}, we enforce \textbf{Exact Match (EM)}. Our preliminary analysis showed that semantic similarity metrics (e.g., BERTScore) fail to penalize subtle scientific errors, assigning high scores ($>0.85$) to plausible yet factually inverted confabulations.

\begin{table*}[t] 
\centering
\small
\setlength{\tabcolsep}{3.5pt}
\renewcommand{\arraystretch}{1.15}

\begin{tabular}{@{}l*{6}{c}@{}}
\toprule
\textbf{Model} & 
\makecell[c]{\textbf{Ind. Judg.}\\(Acc / F$_1$)} & 
\makecell[c]{\textbf{Comp. Judg.}\\(Acc / F$_1$)} & 
\makecell[c]{\textbf{Neg. Loc.}\\(Acc / IoU)} & 
\makecell[c]{\textbf{Ent. Loc.}\\(Acc / IoU)} & 
\makecell[c]{\textbf{Ent. Corr.}\\(Acc)} & 
\makecell[c]{\textbf{Avg.}\\\textbf{Acc.}}\\
\midrule

Gemma-3-1B    & 0.51 / 0.53 & 0.53 / 0.34 & 0.13 / 0.13 & 0.10 / 0.10 & 0.00 & 0.25 \\
Gemma-3-4B    & 0.58 / 0.60 & 0.52 / 0.33 & 0.36 / 0.33 & 0.38 / 0.25 & 0.10 & 0.39 \\
Gemma-3-12B   & 0.65 / 0.63 & \textbf{0.71} / 0.48 & 0.44 / 0.44 & 0.46 / 0.26 & 0.19 & 0.49 \\
Gemma-3-27B   & \textbf{0.71} / 0.72 & 0.56 / 0.54 & 0.61 / 0.46 & 0.46 / 0.29 & 0.24 & 0.52 \\
\addlinespace[0.4em]

Llama-3.2-1B  & 0.49 / 0.39 & 0.50 / 0.27 & 0.00 / 0.00 & 0.08 / 0.04 & 0.01 & 0.22 \\
Llama-3.2-3B  & 0.52 / 0.34 & 0.48 / 0.29 & 0.02 / 0.02 & 0.04 / 0.08 & 0.02 & 0.22 \\
Llama-3.3-70B & 0.67 / \textbf{0.73} & 0.50 / 0.39 & \textbf{0.69} / \textbf{0.61} & 0.13 / 0.24 & 0.16 & 0.43 \\
\addlinespace[0.4em]

GPT-4o-mini   & 0.62 / 0.52 & 0.59 / \textbf{0.55} & 0.59 / 0.54 & 0.07 / 0.20 & 0.21 & 0.42 \\
GPT-4o        & 0.67 / 0.67 & 0.60 / 0.53 & 0.66 / 0.57 & \textbf{0.47} / \textbf{0.38} & \textbf{0.28} & \textbf{0.54} \\
\bottomrule
\end{tabular}
\caption{
Metrics are reported as Accuracy / F$_1$ or Accuracy / IoU, depending on the task. The final column shows the average of \textbf{accuracy} scores only (excluding F$_1$ and IoU). 
\textbf{Bold} indicates the best-performing model in each metric column.
}
\label{tab:benchmark_results}
\end{table*}

\section{Results and Analysis}
\label{sec:results}

\paragraph{General Results}
Table~\ref{tab:benchmark_results} reveals a prevasive struggle across state-of-the-art LLMs on ReFACT. While \texttt{GPT-4o} secures the top performance, its modest aggregate score ($\mathrm{Acc}_{\mathrm{avg}}=0.54$) highlights the difficulty of detecting subtle scientific confabulations. Among open-weight models, \texttt{Gemma-3-27B} (0.52) remarkably outperforms the much larger \texttt{Llama-3.3-70B} (0.43), suggesting that for scientific factuality, data quality or architectural efficiency may outweigh sheer parameter scale.\linebreak 
Critically, all models fail to reliably identify specific errors: even GPT-4o achieves less than 50\% accuracy on entity localization. To explain this systematic deficienty, we investigate the underlying error patterns below.

\subsection{Error Analysis}
\label{sec:error_analysis}
To diagnose failure mechanisms, we categorized 5,826 false positive predictions from the entity localization task relative to the gold span $s^*$. We distinguish seven distinct distractor types:

\begin{itemize}[leftmargin=*]
    \setlength\itemsep{0em}
    \item \textbf{adjacent\_span}: Spatially proximal (overlapping or within 20 chars of $s^*$).
    \item \textbf{partial\_overlap}: Intersects with $s^*$ but misses key tokens.
    \item \textbf{word\_overlap}: Shares surface tokens without semantic alignment.
    \item \textbf{number\_date}: Type mismatch (numerical/temporal) against a non-numerical $s^*$.
    \item \textbf{common\_noun}: Generalization error (generic noun vs. specific entity).
    \item \textbf{long\_fragment}: Over-extraction ($>10$ tokens) containing $s^*$ plus noise.
    \item \textbf{unrelated}: Semantically disconnected from $s^*$.
\end{itemize}

The \textbf{unrelated} category is further subdivided into: (1) stopwords, (2) named entities, (3) technical terms, and (4) general content.

\begin{figure*}[!t]
    \centering
    \includegraphics[width=\textwidth]{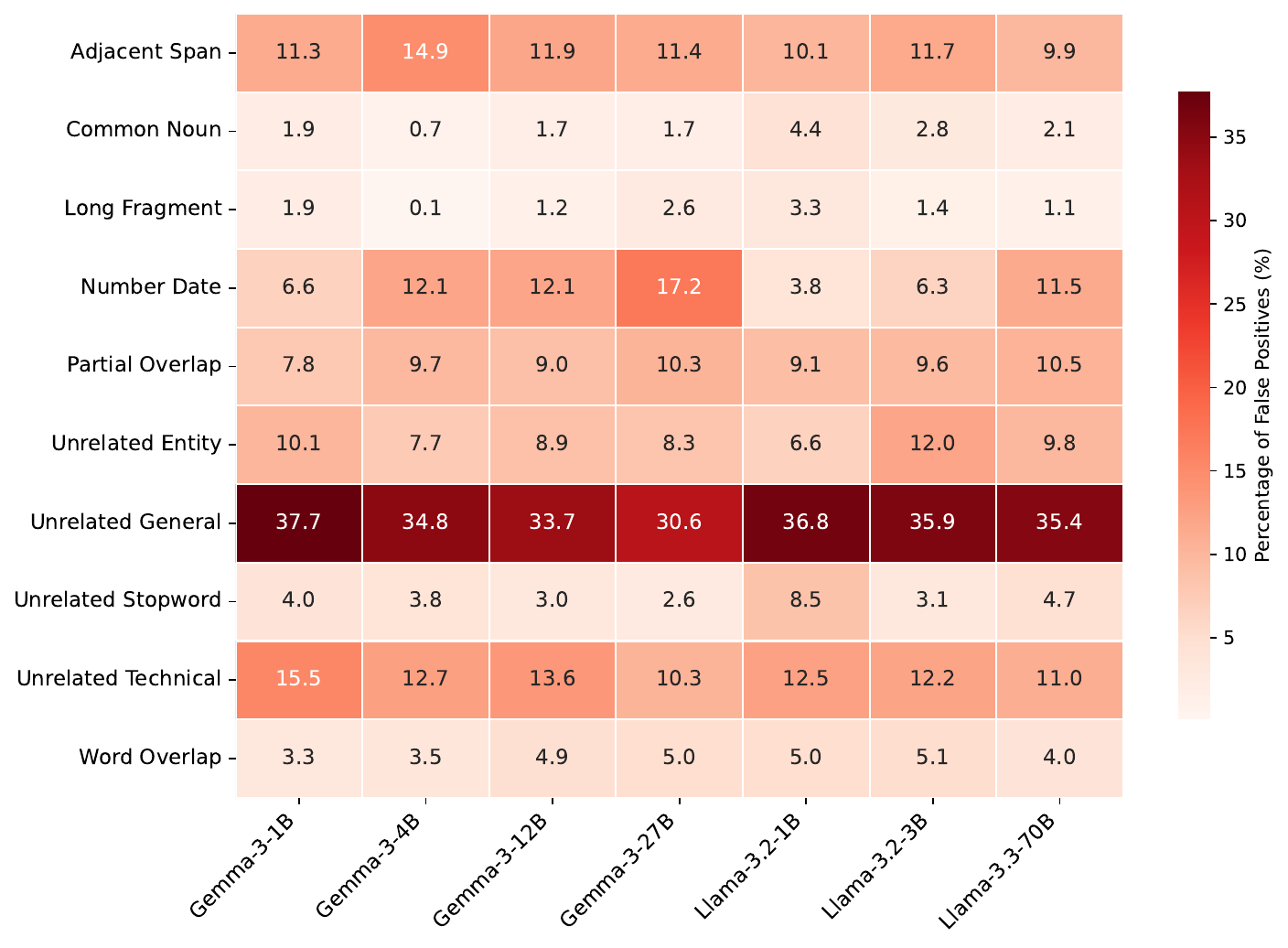}
    \caption{\textbf{The Salient Distractor Phenomenon.} Analysis of 5,826 false positives reveals a systematic failure in semantic grounding: \textbf{61\% of localization errors} focus on contextually salient but semantically unrelated terms, rather than actual factual errors. Crucially, this pattern is \textbf{scale-invariant}, persisting from 1B to 70B models. This confirms that current LLMs rely on surface-level saliency heuristics rather than true factuality detection. For instance, when models fail to detect the error in ``DNA replication'', they often wrongly highlight ``RNA'' simply because it is a biological term, ignoring the actual semantic contradiction.}
    \label{fig:distractor_heatmp}
\end{figure*}

\subsubsection{The Salient Distractor Problem}
\label{sec:salient_distractor}

\paragraph{Semantic Disconnect.}
Our analysis of 5,826 false positive predictions exposes a critical failure mode: \textbf{51.8--67.3\%} of localization errors are semantically unrelated to the actual factual contradiction (\autoref{fig:distractor_heatmp}). Instead of identifying errors, models systematically fixate on \textbf{salient distractors}-contextually plausible technical terms (12.5\%), named entities (8.8\%), or generic scientific phrases (35.1\%). Crucially, this is not a boundary detection issue but a foundamental grounding failure. IoU analysis reveals a stark \textbf{bimodal distribution}: 90.6\% of these false positives share \emph{exactly zero} character overlap with the gold span. Models are not imprecisely locating the error; they are confidently pointing to entirely wrong regions driven by surface-level keyword prominence.

\paragraph{Invariance to Scale and Prompting.}
This semantic deficit proves remarkably resistant to standard scaling and prompting interventions:
\begin{itemize}[leftmargin=1em, nosep]
    \item \textbf{Scaling Limitation:} While unrelated errors decrease slightly from 1B (67\%) to 70B (61\%) models, even \texttt{Llama-3.3-70B} remains dominated by this failure mode. Larger models merely become better at selecting "scientific-sounding" distractors without improving factual discrimination.
    \item \textbf{Prompting Failure:} As shown in \autoref{tab:prompting_ablation}, advanced strategies fail to mitigate this issue. \textbf{Zero-shot prompting consistently outperforms} Few-Shot and Chain-of-Thought (CoT) approaches across all scales (e.g., \texttt{Gemma-3-27B} drops from 0.71 F$_1$ zero-shot to 0.64 with CoT).
\end{itemize}

This counter-intuitive finding suggests that the bottleneck is not instruction adherence or reasoning depth, but the underlying representation of scientific truth.

\paragraph{Transformation Agnosticism.}
Finally, error patterns remain consistent across transformation types (Negation: 62.6\% unrelated vs. Entity Swap: 61.0\%), indicating that the failure is agnostic to the logical complexity of the error. However, subtle nuances exist: Entity Swaps elicit more unrelated entity predictions (10.3\% vs. 6.4\%) as models grasp for alternative nouns, while Negations show higher partial overlap (20.1\%), suggesting models can vaguely sense the region of logical reversal but lack the precision to pinpoint the operator.

\begin{table}[!ht]
\centering
\small
\begin{tabular}{lrrr}
\toprule
Model & Zero-Shot & Few-Shot & Few-Shot + CoT \\
\midrule
Gemma-3-1B & 0.504 & 0.455 & 0.432 \\
Gemma-3-4B & 0.584 & 0.561 & 0.517 \\
Gemma-3-12B & 0.653 & 0.629 & 0.612 \\
Gemma-3-27B & 0.707 & 0.704 & 0.640 \\
Llama-3.2-1B & 0.475 & 0.416 & 0.463 \\
Llama-3.2-3B & 0.481 & 0.412 & 0.487 \\
Llama-3.3-70B & 0.651 & 0.653 & 0.644 \\
\bottomrule
\end{tabular}
\caption{\textbf{Prompting Strategy Ablation for Independent Judgment.} F$_1$ scores across prompting techniques. Zero-shot consistently matches or outperforms few-shot and chain-of-thought prompting, indicating the task difficulty stems from semantic grounding limitations rather than instruction-following challenges.}
\label{tab:prompting_ablation}
\end{table}

\subsection{Task-Specific Performance Analysis}

\paragraph{Confabulation Judgment}
Detecting confabulations remains challenging even for state-of-the-art models. As shown in~\autoref{tab:benchmark_results}, smaller models (Llama-3.2-1B: 0.49, Gemma-3-1B: 0.51) perform near random chance (0.50) on Independent Judgment. Larger models show improvement: Llama-3.3-70B achieves 0.67/0.73 (Acc/F$_1$) and Gemma-3-27B reaches 0.71/0.72, but still fall short of reliable detection.

Contrary to expectations, Comparative Judgment appears to be even more difficult for most models. Across the board, models exhibit lower F$_1$ scores on Comparative Judgment than on Independent Judgment, despite the task involving a relative comparison between two answers. For example, Llama-3.3-70B drops from 0.73 to 0.39 in F$_1$, and GPT-4o drops from 0.67 to 0.53. This suggests that when both answers are plausible yet subtly different, models often struggle to discern which one contains a confabulation, potentially due to shallow comparative heuristics or overconfidence in surface-level coherence. Detailed precision, recall, and F$_1$ scores on original and confabulated answers are given in \autoref{tab:merged_judgment_metrics}.

\paragraph{Implications for LLM-as-Judge Paradigms}
These findings challenge the widespread use of LLMs as evaluators in research. Many recent benchmarks employ GPT-4o or similar models to assess other systems, assuming reliable error detection capabilities. However, our results show that even GPT-4o achieves only 0.60 accuracy on Comparative Judgment, barely above random for binary classification. This poor performance, particularly in scientific contexts requiring domain expertise, suggests LLM-as-Judge paradigms may produce unreliable evaluations for factuality and reasoning tasks. Researchers should reconsider this evaluation methodology, or, at a minimum, report confidence intervals and baseline comparisons.

\paragraph{Confabulation Localization}
The localization results show that LLMs struggle to accurately identify the span that contains a confabulation, particularly in the case of entity replacements. Across all models, performance on negation localization is substantially higher than on entity localization. For example, GPT-4o achieves 0.66 accuracy and 0.57 IoU for negation, whereas its performance drops to 0.47/0.38 for entity replacements. Similarly, Llama-3.3-70B achieves a respectable 0.69/0.61 on negation but only 0.13/0.24 on entity localization, with smaller models often scoring to zero.

This disparity highlights a key challenge: While negated claims may be easier to catch due to their syntactic structures, identifying subtle entity substitutions, especially technical or domain-specific ones, requires deeper semantic understanding that most models seem to lack. Among open-weight models, Gemma-3-27B shows the strongest performance on entity localization (0.46/0.29), yet still falls short of what would be considered robust detection.

\paragraph{Correction} \label{result: correction}
The correction task remains the most challenging task for all evaluated models. Even the best-performing model, GPT-4o, achieves only 0.28 accuracy, far from indicating reliable correction ability. Among open-weight models, Gemma-3-27B model shows the strongest performance at 0.44, while Llama-3.3-70B falls behind at 0.23. Smaller models perform drastically worse, for instance, Llama-3.2-1B achieves only 0.05, and Gemma-3-1B drops to 0.03. 

Although we computed BERTScore values, we omit them from our  primary results. This metric often yields inflated scores for factually incorrect predictions---e.g., BERTScore exceeds 0.85 for multiple models despite very low EM accuracy, making it unsuitable for evaluating factual correction.

\section{Conclusion}
\label{sec:conclusion}

We identify \textbf{salient distractor selection} as a dominant, scale-invariant failure mode in LLM confabulation detection. Across 5,826 false positives, 61\% target semantically unrelated spans—90.6\% with zero character overlap to actual errors. This pattern persists from 1B to 70B parameters (67\% $\rightarrow$ 61\% unrelated), across transformation types (negation: 62.6\%, entity-swap: 61.0\%), and resists prompting interventions, exposing a fundamental semantic grounding deficit that scaling alone cannot resolve.

\textbf{Comparative judgment proves paradoxically harder than independent detection}: GPT-4o's F$_1$ drops from 0.67 to 0.53 when evaluating answers side-by-side. Entity errors pose particular difficulty—localization accuracy falls to 0.47 versus 0.66 for negations, and correction reaches only 0.28 even for GPT-4o. These results directly challenge LLM-as-Judge paradigms for scientific factuality.

Future work should target these failure modes through contrastive training with hard negatives, hierarchical localization objectives, and explicit entity grounding mechanisms. We aim to extend ReFACT across domains to deepen understanding of LLM factuality.

\section*{Limitations}
\paragraph{Data Constraints \& Scope.} 
Our benchmark relies on community-curated scientific discourse (r/AskScience). While we mitigate noise by filtering for high-consensus answers, latent crowd-sourced inaccuracies may remain. Furthermore, we strictly target fine-grained, \textit{span-level} confabulations; this design offers precise error localization but does not address broader, document-level thematic contradictions.

\paragraph{Annotation Ambiguity.}
Distinguishing subtle confabulations from facts entails inherent subjectivity. We address this by enforcing a three-way annotator consensus protocol, prioritizing high inter-annotator agreement over dataset scale.

\paragraph{Evaluation Bias \& Metrics.}
We acknowledge potential \textit{self-preference bias} when models evaluate their own generations. To minimize this, we maximize architectural diversity (e.g., Llama, GPT, Gemma) between generation and evaluation stages. Finally, as automatic metrics remain imperfect proxies for human judgment, we interpret results as relative performance trends rather than absolute measures of truth.

\section*{Ethics Statement}
\paragraph{Selection Bias \& Content Quality.}
We acknowledge inherent selection biases in sourcing from Reddit (r/AskScience), where visibility correlates with popularity rather than purely scientific merit. To mitigate this, our annotation guidelines strictly filter for objective scientific validity, excluding anecdotal, humorous, or institutionally-biased responses. While we employ automated and human filtering to remove toxic or inappropriate language, the dataset reflects the linguistic characteristics of public internet discourse.

\paragraph{Privacy \& Data Compliance.}
Data collection utilized the public Pushshift API, consistent with Reddit's Terms of Service. As individual consent is infeasible at scale, we adhere to a strict non-identification policy: all user identifiers were removed, and content was screened to eliminate Personal Identifiable Information (PII). The released dataset contains exclusively scientific explanations, posing minimal risk to original content creators.

\section*{Acknowledgments}

This research was funded in part by the German Federal Ministry of Research, Technology and Space under the funding code “KI-Servicezentrum Berlin-Brandenburg” 16IS22092. Responsibility for the content remains with the authors.

\bibliography{anthology, custom}

\clearpage
\appendix

\lstset{
    basicstyle=\small,
    breaklines=true,
    postbreak=\mbox{\textcolor{gray}{$\hookrightarrow$}\space},
    breakindent=1em,
}

\begin{table*}[t]
\label{sec:detailed_judgement_result}
\centering
\small
\renewcommand{\arraystretch}{1.15}
\setlength{\tabcolsep}{3.5pt}

\begin{tabular}{@{}l cccc c cccc@{}}
\toprule
& \multicolumn{4}{c}{\textbf{Independent Judgment}} && \multicolumn{4}{c}{\textbf{Comparative Judgment}} \\
\cmidrule(lr){2-5} \cmidrule(lr){7-10}
\textbf{Model} & 
\textbf{Acc} & \textbf{Prec} & \textbf{Rec} & \textbf{F$_1$ (C/O)} && 
\textbf{Acc} & \textbf{Prec} & \textbf{Rec} & \textbf{F$_1$ (C/O)} \\
\midrule
Gemma-3-1B    & 0.51 & 0.51 & 0.56 & 0.53 / 0.48 && 0.53 & 0.36 & 0.35 & 0.34 / 0.34 \\
Gemma-3-4B    & 0.58 & 0.58 & 0.62 & 0.60 / 0.57 && 0.52 & 0.37 & 0.36 & 0.33 / 0.33 \\
Gemma-3-12B   & 0.65 & \textbf{0.68} & 0.58 & 0.63 / \textbf{0.68} && \textbf{0.71} & 0.48 & 0.48 & 0.48 / 0.48 \\
Gemma-3-27B   & \textbf{0.71} & 0.69 & 0.75 & 0.72 / 0.69 && 0.56 & \textbf{0.55} & \textbf{0.55} & 0.54 / 0.54 \\
\addlinespace[0.4em]

Llama-3.2-1B  & 0.49 & 0.48 & 0.33 & 0.39 / 0.56 && 0.50 & 0.32 & 0.33 & 0.27 / 0.27 \\
Llama-3.2-3B  & 0.52 & 0.54 & 0.25 & 0.34 / 0.62 && 0.48 & 0.33 & 0.32 & 0.29 / 0.29 \\
Llama-3.3-70B & 0.67 & 0.62 & \textbf{0.90} & \textbf{0.73} / 0.57 && 0.50 & 0.52 & 0.50 & 0.39 / 0.39 \\
\addlinespace[0.4em]

GPT-4o-mini   & 0.62 & 0.70 & 0.41 & 0.52 / 0.68 && 0.59 & 0.65 & 0.59 & \textbf{0.55} / \textbf{0.55} \\
GPT-4o        & 0.67 & 0.67 & 0.66 & 0.67 / 0.67 && 0.60 & \textbf{0.73} & 0.59 & 0.53 / 0.53 \\
\bottomrule
\end{tabular}

\caption{
\textbf{Detailed Judgment Metrics.} Evaluation of Independent vs. Comparative Judgment tasks. We report Accuracy (Acc), Precision (Prec), Recall (Rec), and F$_1$ scores. For F$_1$, we report values for both the Confabulated (C) and Original (O) classes (format: C / O). \textbf{Bold} indicates the best performance per column.
}
\label{tab:merged_judgment_metrics}
\end{table*}

\section{Transformation Prompts}
\label{sec:transformation_prompts}
For reproducibility, we list the prompts used for the transformations described in \autoref{sec:dataset_creation_pipeline}.

\subsection{Facts Extraction}

\small
\noindent \texttt{Extract all facts mentioned in the following text that are not opinions or beliefs. Be as concise as possible. If there are no factual statements, return ``<bad-example>'' Otherwise, return the facts enclosed by ``<fact>'' and ``</fact>''. Don't add any explanations}

\subsection{Negation}
\small 
\noindent \texttt{Given a paragraph and a specific sentence from the paragraph, negate that sentence. The negation shall not contradict the paragraph but be factually wrong.
Negate ONLY the specific sentence. Output ONLY the negated sentence. No Explanation.}

\subsection{Replacement: Select Keyphrase}

\small 
\noindent \texttt{Return a key phrase from the input text, which is not part of the question. This key phrase should:}

\begin{itemize}[leftmargin=*, itemsep=-1.5mm]
    \item \texttt{Contain factual information or be a technical term.}
    \item \texttt{Be hard to detect by individuals not educated in the domain, preferably specialized terms or phrases less familiar to the general public.}
    \item \texttt{Not be inside of the question.}
\end{itemize}
\texttt{Only return the original key phrase from the input text in list form. Do not provide any explanation. \\}

\noindent \texttt{Example 1:\\
    Question: Where is the Reichstag?\\
    Input Text: The Reichstag is in Berlin, Germany.\\
    Output: [``Berlin'']}
\vspace{0.2cm}

\noindent \texttt{Example 2:\\
    Question: Could you name me a deadly virus?  \\        
    Input Text: Covid-19 is a deadly virus. To the public, it is more commonly known as Corona or SARS-CoV-2.\\
    Output: [``SARS-CoV-2'']}
\vspace{0.2cm}

\noindent \texttt{Example 3:\\
    Question: What is the Hasso-Plattner-Institute in Potsdam called? \\ 
    Input Text: It is called Hasso-Plattner-Institute. The Hasso-Plattner-Institute was founded by Hasso Plattner.\\
    Output: [``Hasso Plattner'']}
\vspace{0.2cm}

\noindent \texttt{Question:\{question\_context\}\\
Input Text: \{text\}\\
Output: }

\subsection{Replacement: Coreference Resolution}

\small 
\noindent \texttt{You are a Coreference Resolution detector. Detect where the exact keyphrase is referred to as the same entity in the input text, excluding pronouns.\\
Instructions:}

\begin{itemize}[leftmargin=*, itemsep=-1.5mm]
    \item \texttt{Include only exact matches or direct references to the keyphrase within the context.}
    \item \texttt{Do not include unrelated mentions or vague references.}
    \item \texttt{Output should list parts of the text where the entity is mentioned again, maintaining the same context.}
    \item \texttt{Do not include any pronouns.}
\end{itemize}
\noindent \texttt{Output: List format without explanations.}
        
\vspace{0.2cm}

\noindent \texttt{Example 1\\ 
      Input Text: Covid 19 is a deadly virus. To the public it is more known as Corona or Sars Covid.\\
      Input Keyphrase: Corona\\
      Output: [``Covid 19'' , ``Sars Covid'']}
\vspace{0.2cm}

\noindent \texttt{Example 2\\
      Input Text: The Hasso Plattner Institute was founded by Hasso Plattner. He also founded SAP.\\
      Input Keyphrase: Hasso Plattner Institute\\
      Output: ['Hasso Plattner Institute']}
\vspace{0.2cm}

\noindent \texttt{Example 3\\
      Input Text: Alcohol is dangerous for humans. \\
      Input Keyphrase: Alcohol\\
      Output: []}
\vspace{0.2cm}

\noindent \texttt{Example 4\\
      Input Text: The Amazon rainforest is known for its biodiversity.\\
      Input Keyphrase: biodiversity\\
      Output: []}
\vspace{0.2cm}

\noindent \texttt{Example 5\\
      Input Text: My head hurts because of a headache. That is bad.\\
      Input Keyphrase: headache\\
      Output: []}
            
\vspace{0.2cm}

\noindent \texttt{Input Text: {text}\\
Input Keyphrase: {keyphrase}\\
Output:}

\subsection{Replacement: Suggest new Keyphrases}
\small
\noindent \texttt{Create false information by changing the tagged substrings. 
        Ensure the rewritten answer to be coherent, convincing and false. 
        For preprocessing reasons only return the altered substrings in list form. 
        No explanation.}
\vspace{0.2cm}

\noindent \texttt{Example 1:\\
            Input: The <replace>Hasso Plattner Institute</replace> for Digital Engineering gGmbH (German: <replace>Hasso-Plattner-Institut</replace> fuer Digital Engineering gGmbH; <replace>HPI</replace>) is an information technology non-profit company affiliated with the University of Potsdam in Potsdam. The teaching and research of <replace>HPI</replace> are focused on 'IT-Systems Engineering'. <replace>HPI</replace> was founded in 1998, and is the first, and as of 2018, the only entirely privately funded faculty[3] in Germany. It is financed entirely through private funds donated by billionaire <replace>Hasso Plattner</replace>,[4] who co-founded the software company <replace>SAP</replace>, and is currently the chairman of <replace>SAP</replace>'s supervisory board. In addition to Hasso Plattner and Christoph Meinel, the management of <replace>HPI</replace> was expanded to include other board members on 2019.\\
            Tagged Substrings: ['Hasso Plattner Institute', 'Hasso-Plattner-Institut', 'HPI', 'HPI', 'HPI', 'Hasso Plattner', 'SAP', 'SAP', 'HPI']\\
            Output: ['Mark Zuckerberg Institute', 'Mark-Zuckerberg-Institut', 'MZI', 'MZI', 'MZI', 'Mark Zuckerberg', 'Facebook', 'Meta', 'MZI']}
\vspace{0.2cm}

\noindent \texttt{Example 2:\\
            Input: Same reason you can feel someone else's pain when they get hurt, e.g. Guys can feel another guy's pain when he gets kicked in the testes. We have <replace>mirror neurons</replace> in our brain. When you do a certain activity or feel a certain emotion, a network of neurons in your brain fire. Now when you see someone who is doing this activity or going through this emotional experience, a subset of those neurons fire in your brain! This probably evolved so that we can feel empathy and relate to what other people are going through, since this is important for living in large societies like humans do.\\
            Tagged Substrings: ['mirror neurons']\\
            Output: ['echo neurons']}
\vspace{0.2cm}

\noindent \texttt{Example 3:\\
            Input: The circle (it's not a spiral) only occurs above the <replace>two rotational poles</replace>. That circle is caused by the Earth rotating. Those star tracks each occupy about 50 degrees or so, so we can estimate that the photo was taken using about a 3.5 hour exposure. You should be able to get a similar photo anywhere on the planet, although the closer to the <replace>equator</replace> you are, the closer the rotation centre will be to the horizon.\\
            Tagged Substrings: ['two rotational poles', 'equator']\\
            Output: ['equator', 'two rotational poles']}
\vspace{0.2cm}

\noindent \texttt{Example 4:\\
            Input: I'm not quite sure what you mean here. However, the most common way of transmitting COVID-19 is through <replace>the air</replace>.\\
            Tagged Substrings: ['the air']\\
            Output: ['contaminated surfaces']}
\vspace{0.2cm}

\noindent \texttt{Example 5:\\
            Input: Because the stability and binding of the nucleus depends on the neutrons just as much as it does the protons. If you have too many or too few neutrons for a given number of protons, you'll no longer have a bound system. The nucleus will break apart on timescales characteristic of the strong force (10**(-22) seconds). The boundaries between bound nuclei and unbound nuclei described above are called the *<replace>driplines</replace>*.\\
            Tagged Substrings: ['driplines']\\
            Output: ['stabilitylines']}
\vspace{0.2cm}

\noindent \texttt{Input: \{question\_context + tagged\_answer\}\\
Tagged Substrings:\{keyphrases\}\\
Output: }

\subsection{Most Convincing}
\label{prompt:mc}

\small
\noindent \texttt{You will get a list of \{len(texts)\} paragraphs manipulated to be factually incorrect.  
Return the index of the most convincing and coherent paragraph, which is still factually incorrect. 
No explanation. }

\clearpage
\section{Dataset Statistics}
\begin{figure}[ht]
    \centering
    \includegraphics[width=\linewidth]{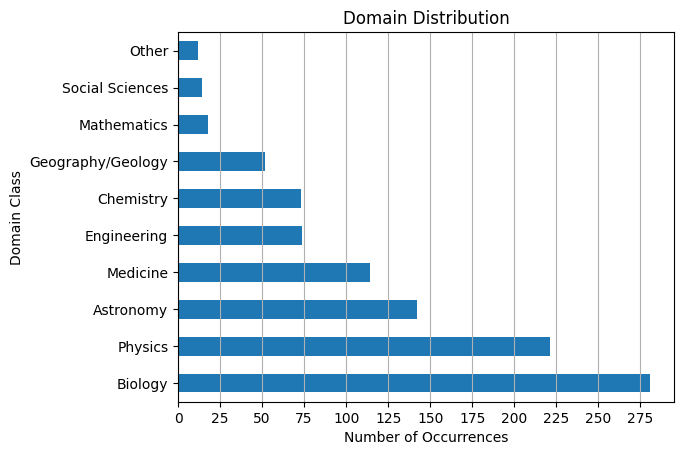}
   \caption{Domains of the Dataset}
    \label{fig:domains}
\end{figure}

\begin{figure}[ht]
    \centering
    \begin{subfigure}[b]{0.9\linewidth}
        \centering
        \includegraphics[width=\linewidth]{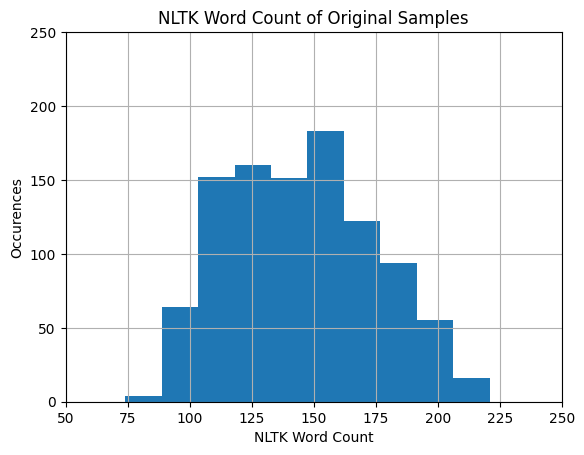}
    \end{subfigure}
    \begin{subfigure}[b]{0.9\linewidth}
        \centering
        \includegraphics[width=\linewidth]{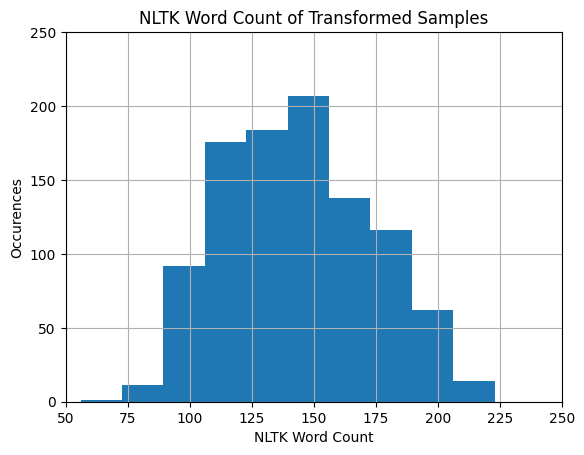}
    \end{subfigure}
    \caption{Word Count of Dataset Samples}
    \label{fig:nltk_count}
\end{figure}

\begin{figure}[ht]
    \centering
    \begin{subfigure}[b]{0.9\linewidth}
        \centering
        \includegraphics[width=\linewidth]{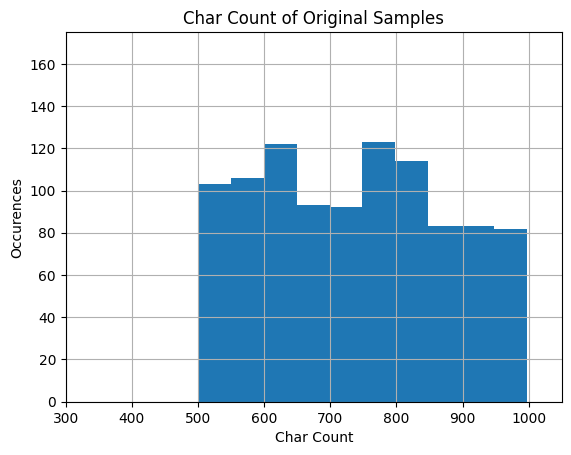}
    \end{subfigure}
    \begin{subfigure}[b]{0.9\linewidth}
        \centering
        \includegraphics[width=\linewidth]{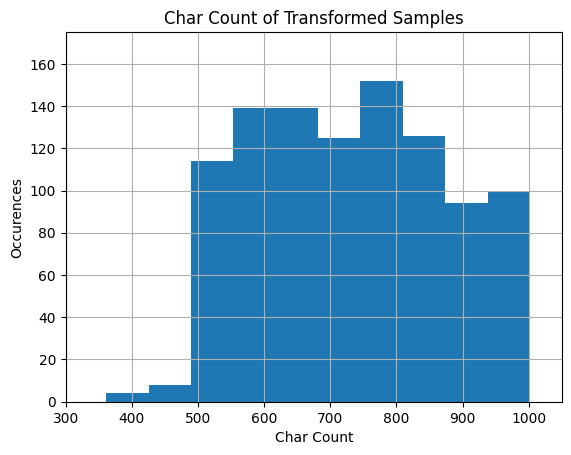}
    \end{subfigure}
    \caption{Character Count of Dataset Samples}
    \label{fig:char_count}
\end{figure}

\begin{figure}[ht]
    \centering
\includegraphics[width=\linewidth]{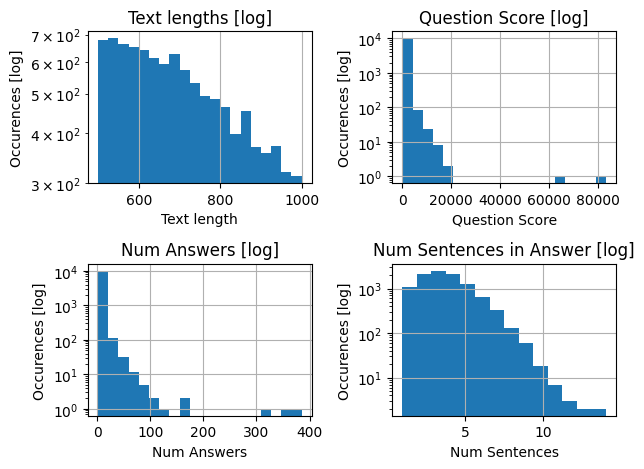}
\caption{Statistics of the r/AskScience sample that was used as a basis for our dataset. }
\label{fig:filtered-dataset-distribution}
\end{figure}

The figures in this section illustrate key dataset statistics, providing insight into its structure and diversity. \autoref{fig:domains} visualizes the distribution of dataset domains, highlighting its thematic range. \autoref{fig:nltk_count} and \autoref{fig:char_count}  plot the distribution of word and character counts before and after the transformation, calculated using the NLTK library for tokenization \cite{nltk}. Lastly, \autoref{fig:filtered-dataset-distribution}  presents statistics for the r/AskScience subset that was used to create the benchmark.

\label{sec:dataset-statistics}

\clearpage
\section{Experiment Prompts}
\label{sec:experiment_prompts}
In the following, we list the prompts used in the experiments described in \autoref{sec:experiments}.

\subsection{Independent Confabulation Judgment}
\label{p:detection_1}
\small
\vspace{0.2cm}
\noindent \textbf{System Prompt:}
\vspace{0.2cm}

\noindent \texttt{Your task is to evaluate the factual correctness of a given answer to a question. Read the entire answer and question carefully, then analyze its factual accuracy step by step. After your analysis, provide a final verdict of either ``True'' if the entire answer is factually correct, or ``False'' if any part of the answer contains inaccuracies or hallucinations.}
\vspace{0.2cm}

\vspace{0.4cm}
\noindent \textbf{User Prompt:}
\vspace{0.2cm}

\noindent \texttt{Task:\\
Question: \{question\} \\
Answer: \{answer\}\\
Final Verdict: }

\subsection{Comparative Confabulation Judgment}
\label{p:detection_3}
\vspace{0.2cm}
\noindent \textbf{System Prompt}
\vspace{0.2cm}

\noindent \texttt{Your task is to return the factually correct answer out of the two given answers (A and B) to a question. Read both entire answers and the question carefully, then analyze the factual accuracy of both answers within the context. After your analysis, provide a final verdict of either answer A or answer B is factually correct.\\}

\noindent \textbf{User Prompt}
\vspace{0.2cm}

\noindent \texttt{Question: \{question\}\\
Answer A: \{answer\_a\}\\
Answer B: \{answer\_b\}\\
Final Verdict: }

\subsection{Negation Confabulation Localization}
\label{p:position_3}
\vspace{0.2cm}
\noindent \textbf{System Prompt}
\vspace{0.2cm}

\noindent \texttt{You will get an answer to a question with one factually wrong sentence inside the answer, which was changed beforehand. Your task is to locate the factually wrong sentence of the fake answer to the question. Read the entire answer with the factually wrong sentence and the corresponding question carefully. Then analyze the factual accuracy of every part in the given answer.  After your analysis, return only the whole sentence without changes.\\}

\noindent \textbf{User Prompt}
\vspace{0.2cm}

\noindent \texttt{Question: \{question\}\\
Answer: \{transformed\_answer\}\\
Wrong Sentence: }

\subsection{Entity Confabulation Localization}
\label{p:position_4}
\vspace{0.2cm}
\noindent \textbf{System Prompt}
\vspace{0.2cm}

\noindent \texttt{You will get an answer to a question with factually wrong entities inside the answer, which were changed beforehand. An entity can be a single word or multiple words of any type. Your task is to locate the factually wrong entities of the fake answer to the question. Read the entire answer with factually wrong entities and the corresponding question carefully. Then analyze the factual accuracy of every part in the given answer with the focus on factually wrong entities. After your analysis, return the factually wrong entities separated with newlines without changes.\\}

\noindent \textbf{User Prompt}
\vspace{0.2cm}

\noindent \texttt{Question: \{question\} \\
Answer: \{transformed\_answer\}\\
Wrong Entities: }

\subsection{Entity Confabulation Correction}
\label{p:fill}

\vspace{0.2cm}
\noindent \textbf{System Prompt}
\vspace{0.2cm}

\noindent \texttt{Your task is to return replacements for the <mask> tags inside an answer to a question. Read the entire answer and question carefully, then analyze the answer and think about possible replacements. After your analysis, return only the list of replacements in the order they appear separated by new line.\\}

\noindent \textbf{User Prompt}
\vspace{0.2cm}

\noindent \texttt{Task:\\
Question: \{question\}\\
Answer: \{answer\}\\
\{number\_of\_masks\} Replacements expected\\
Replacements: }

\end{document}